\documentclass[preprint,12pt]{elsarticle}

\usepackage{amssymb}
\usepackage{amssymb}
\usepackage{float}
\usepackage {amsmath}
\usepackage{algorithm}  
\usepackage{algpseudocode}  
\usepackage{amsmath}  
\usepackage{booktabs} 
\usepackage{caption} 
\usepackage{setspace}
\usepackage{subcaption}
\usepackage{hyperref}
\usepackage{threeparttable}
\usepackage{multirow}
\usepackage{doi}
\usepackage{xcolor} 
\usepackage{caption}
\usepackage{mathrsfs}
\usepackage{bm}
\usepackage{adjustbox}

\hypersetup{
    colorlinks=true, 
    linkcolor=blue,  
    urlcolor=blue,   
    citecolor=blue   
}

\makeatletter
\renewcommand{\fnum@figure}{\textbf{Fig. \thefigure.}\@gobble}
\makeatother

\makeatletter
\renewcommand{\fnum@table}{\textbf{Table \thetable }\@gobble}
\makeatother

\journal{Elsevier}

\begin{document}

\begin{frontmatter}

\title{Attention-Based Multiscale Temporal Fusion Network for Uncertain-Mode Fault Diagnosis in Multimode Processes}

\author[a]{Guangqiang Li} 
\ead{guangqiangli@whut.edu.cn}

\author[b]{M. Amine Atoui} 
\ead{amine.atoui@gmail.com}

\author[a]{Xiangshun Li\corref{cor1}} 
\ead{lixiangshun@whut.edu.cn}
\cortext[cor1]{Corresponding author.}

\affiliation[a]{organization={School of Automation},
            addressline={Wuhan University of Technology}, 
            city={Wuhan},
            postcode={430070}, 
            country={PR China}}
\affiliation[b]{organization={The School of Information Technology},
            addressline={Halmstad University}, 
            city={Halmstad},
            country={Sweden}}

\begin{abstract}
Fault diagnosis in multimode processes plays a critical role in ensuring the safe operation of industrial systems across multiple modes.
It faces a great challenge yet to be addressed – that is, 
the significant distributional differences among monitoring data from multiple modes make it difficult for the models to extract shared feature representations related to system health conditions.
In response to this problem, this paper introduces a novel method called attention-based multiscale temporal fusion network.
The multiscale depthwise convolution and gated recurrent unit are employed to extract multiscale contextual local features and long-short-term features.
Instance normalization is applied to suppress mode-specific information.
Furthermore, a temporal attention mechanism is designed to focus on critical time points with higher cross-mode shared information, thereby enhancing the accuracy of fault diagnosis.
The proposed model is applied to Tennessee Eastman process dataset and three-phase flow facility dataset. 
The experiments demonstrate that the proposed model achieves superior diagnostic performance and maintains a small model size.
The source code will be available on GitHub at \href{https://github.com/GuangqiangLi/AMTFNet}{https://github.com/GuangqiangLi/AMTFNet}.

\end{abstract}



\begin{keyword}
Fault diagnosis \sep Multimode process \sep Deep learning
\end{keyword}

\end{frontmatter}

\section{Introduction}
\label{sec:introduction}


Safety has always been the primary concern in the operation of industrial systems.
As the scale and complexity of modern industrial processes continue to increase, both the probability of faults and their potential impact have grown significantly \cite{RN266414,RN260370,RN266415}. 
Faults are typically characterized by one or more system parameters deviating from acceptable operating ranges \cite{RN266413}. 
It may lead to severe incidents, resulting in significant economic losses, environmental damage and threats to human safety \cite{RN246049,RN266416}. 
Thus, as an important tool for assessing the system health condition, fault diagnosis plays an essential role in ensuring the safe and reliable operation of industrial systems.

With the deployment of a large number of sensors, industrial systems collected massive amounts of monitoring data, and there is a growing interest in data-driven fault diagnosis (DDFD) efforts \cite{RN260374}.
Traditional data-driven methods such as principal component analysis (PCA) \cite{RN260377}, independent component analysis (ICA) \cite{RN260378}, support vector machines (SVM) \cite{RN260382}, and Bayesian network (BN) \cite{RN260384} have been extensively studied and applied in fault diagnosis. 
To further enhance diagnostic performance, various hybrid methods have been proposed, including the combination of PCA and BN \cite{RN266418,atoui2020fault}, the combination of naive Bayes classifier with BN and event tree analysis \cite{RN266419}, and the combination of R-vine copula with event tree \cite{RN266420}. 
The increasing complexity of modern industrial systems imposes higher requirements on the accuracy and intelligence of fault diagnosis methods. 
In this context, deep learning methods have gained significant attention for their ability to automatically extract deep features and provide high diagnostic accuracy \cite{RN260385}.
Wu et al. \cite{RN260387} used the convolutional neural network (CNN) to extract the feature representation. 
Liu et al. \cite{RN260390} integrated the sparse autoencoder with the denoising autoencoder to capture robust, sparse but intrinsic nonlinear features.  
Huang et al. added the long short-term memory (LSTM) network to CNN to capture temporal features \cite{RN260391}.
Zhou et al. employed the self-attention mechanism to learn global information \cite{RN260425}.
Zhu et al. combined multiscale and bidirectional mechanism for feature extraction \cite{RN260405}.
These models primarily focus on fault diagnosis under a single operating mode.
However, as the environmental conditions, loads, and production schedules change, industrial systems often switch between multiple operating modes. 
The mode switching leads to complexity in data distribution, which weakens the performance of existing fault diagnosis methods.

Recently, some studies have focused on fault diagnosis in multimode processes.
These works primarily address the challenge of distribution differences between the training data (source domain) and the test data (target domain).
This scenario is referred to as cross-domain fault diagnosis.
The techniques employed in cross-domain fault diagnosis methods mainly include data extension and representation learning.
Data extension promotes the learning of cross-domain generalizable feature representations by improving the variety of training samples. 
The methods for increasing data diversity include data augmentation techniques (e.g., MixUp \cite{RN260435} and data transformations \cite{RN260434}) and data generation via deep neural networks \cite{RN260436,RN260437}.
However, generating high-quality samples is difficult.
Representation learning mainly focuses on reducing the variation between source and target domains. 
These methods include adversarial-based methods and metric-based methods.
In adversarial-based methods, adversarial learning is employed to deceive the discriminator, thereby capturing the domain-invariant features \cite{RN260398,RN260400,RN260402,RN260439,RN260440,RN262328}.
Metric-based methods explicitly align features from different domains to capture generalizable feature representations. 
This is typically achieved by incorporating distance metrics into the loss function \cite{RN265754,RN260394,RN260395,RN260397,RN260403,RN260430,RN260431}.

Although these fault diagnosis methods for multimode processes have achieved significant progress, some issues are still not well addressed. 
On the one hand, current research focuses on fault diagnosis in multimode processes by capturing domain-invariant features. 
However, the varying amount of domain-invariant information contained in different time points is often neglected. 
Under the influence of faults and control loops, the system undergoes a transient-to-steady evolution. 
Some time steps may show behaviors inconsistent with the final steady-state impact due to control compensation or system inertia. 
Uniformly using all time steps may introduce irrelevant or misleading features. Focusing on critical moments helps capture fault responses that are less affected by control, thus containing more domain-invariant features.
And overemphasizing time points with limited domain-invariant information can decrease the success of fault diagnosis.
On the other hand, the operating mode of the samples is typically inaccessible, which makes methods relying on known operating modes inapplicable.

To address the above problems, a model named attention-based multiscale temporal fusion network (AMTFNet) is proposed. 
Specifically, the multiscale depthwise convolution (MSDC) and gated recurrent unit (GRU) are employed to capture multiscale contextual local features and long-short-term feature representations, respectively. 
Additionally, a temporal attention mechanism (TAM) is constructed to assign weights to deep features at different time steps, thereby enhancing the importance of specific time features that contain more domain-invariant information.
The primary contributions of this study are outlined below.

(1) An attention-based multi-scale temporal fusion network (AMTFNet) is proposed for uncertain-mode fault diagnosis. The model enables accurate diagnosis of multimode processes without requiring prior knowledge of the specific operating mode. 

(2) Under the combined influence of faults and control loops, the system exhibits dynamic behaviors that may be inconsistent with the final steady-state response, potentially introducing misleading features. To address this problem, a temporal attention mechanism is designed to focus on critical time steps with more domain-invariant fault information, thereby enhancing diagnostic performance under diverse operating modes.

(3) Compared to recent advanced methods that do not explicitly address the uncertain-mode scenario considered here, AMTFNet achieves effective fault diagnosis in this novel setting while maintaining a compact model size. 

The subsequent sections are arranged as follows.
In \hyperref[Sec2]{Section \ref{Sec2}}, the problem description, related research areas and preliminary theoretical knowledge of depthwise convolution and GRU are introduced.
The in-depth explanation of the proposed method is presented in \hyperref[Sec3]{Section \ref{Sec3}}.
In \hyperref[Sec4]{Section \ref{Sec4}}, the experiments are conducted on the two datasets to demonstrate the performance of the proposed method.
The final conclusions are summarized in \hyperref[Sec5]{Section \ref{Sec5}}.

\section{Preliminaries}
\label{Sec2}
\subsection{Problem formulation}

Industrial systems switch between multiple stable modes as environments and production schedules change. 
Assume that a industrial system is designed to operate in $M$ modes, and the monitoring data collected from these modes is denoted as $D=\{(\bm{x}_{k}^\text{o},y_k)\}_{k=1}^{N}\sim P_{XY}$, where $N$ denotes the number of samples corresponding to the $M$ modes. 
Here, $\bm{x}_{k}^\text{o} \in \mathbb{R}^v$ denotes the monitoring data at the $k$-th time point, with $v$ being the dimension of the measured variables, and $y_k\in\{1,...,L\}$ denotes the system health condition labels, including one normal state and $L-1$ fault states.
Although it is known that the samples are collected from $M$ modes, the specific mode to which each sample belongs is unknown.
In uncertain-mode fault diagnosis, it is assumed that fault data for the $M$ modes have been collected, with each mode containing samples for $L$ health conditions. 
The goal of uncertain-mode fault diagnosis is to construct a model trained on the monitoring data of these $M$ modes, such that the health condition of the system in these modes can be effectively identified.

\subsection{Related research areas}
There are several research fields related to uncertain-mode fault diagnosis (UMFD), including but not limited to: single mode fault diagnosis (SMFD), domain adaptation-based fault diagnosis (DAFD), and domain generalization-based fault diagnosis (DGFD). The comparison between these methods and UMFD is presented in \hyperref[Table1]{Table 1}.

\begin{table}[!ht]
\centering
	\label{Table1}
    \fontsize{10}{14}\selectfont
    	\begin{threeparttable}
		\caption{Comparison between UMFD and some related learning paradigms.}
        
		\begin{tabular}{llll}
\hline
\textbf{Settings} & \textbf{Training set}                                                    & \textbf{Testing set}    & \textbf{\begin{tabular}[c]{@{}l@{}}Mode availability \\ of the $k$-th sample $r_k$\end{tabular}} \\ \hline
\textbf{SMFD}     & $D^i$                                                                    & $D^{i}$                 & $r_k=i$                                                                                    \\
\textbf{DAFD}     & $\{ D^i \}_{i=1}^S\cup\{ D^{’j} \}_{j=S+1}^M$ \tnote{*} & $\{ D^{j} \}_{j=S+1}^M$ & $r_k \in \{1,…,M\} \text{ and } r_k=r_0$                                                     \\
\textbf{DGFD}     & $\{ D^i \}_{i=1}^S$                                                      & $\{ D^{j} \}_{j=S+1}^M$ & $r_k \in \{1,…,M\} \text{ and } r_k=r_0$                                                     \\
\textbf{UMFD}     & $\{ D^i \}_{i=1}^M$                                                      & $\{ D^{i} \}_{i=1}^M$   & $r_k \in \{1,…,M\}$                                                                        \\ \hline
		\end{tabular}
		\begin{tablenotes}
			\footnotesize
			\item[*] {$D’^{j}$ denotes the monitoring data without system health condition label. $D’^{j}=\{(\bm{x}_k^j)\}_{k=1}^{N_j}$.}
		\end{tablenotes}
	\end{threeparttable}

\end{table}

\begin{figure}[!ht]
\centerline{\includegraphics[width=1.\columnwidth,height=0.6595\columnwidth]{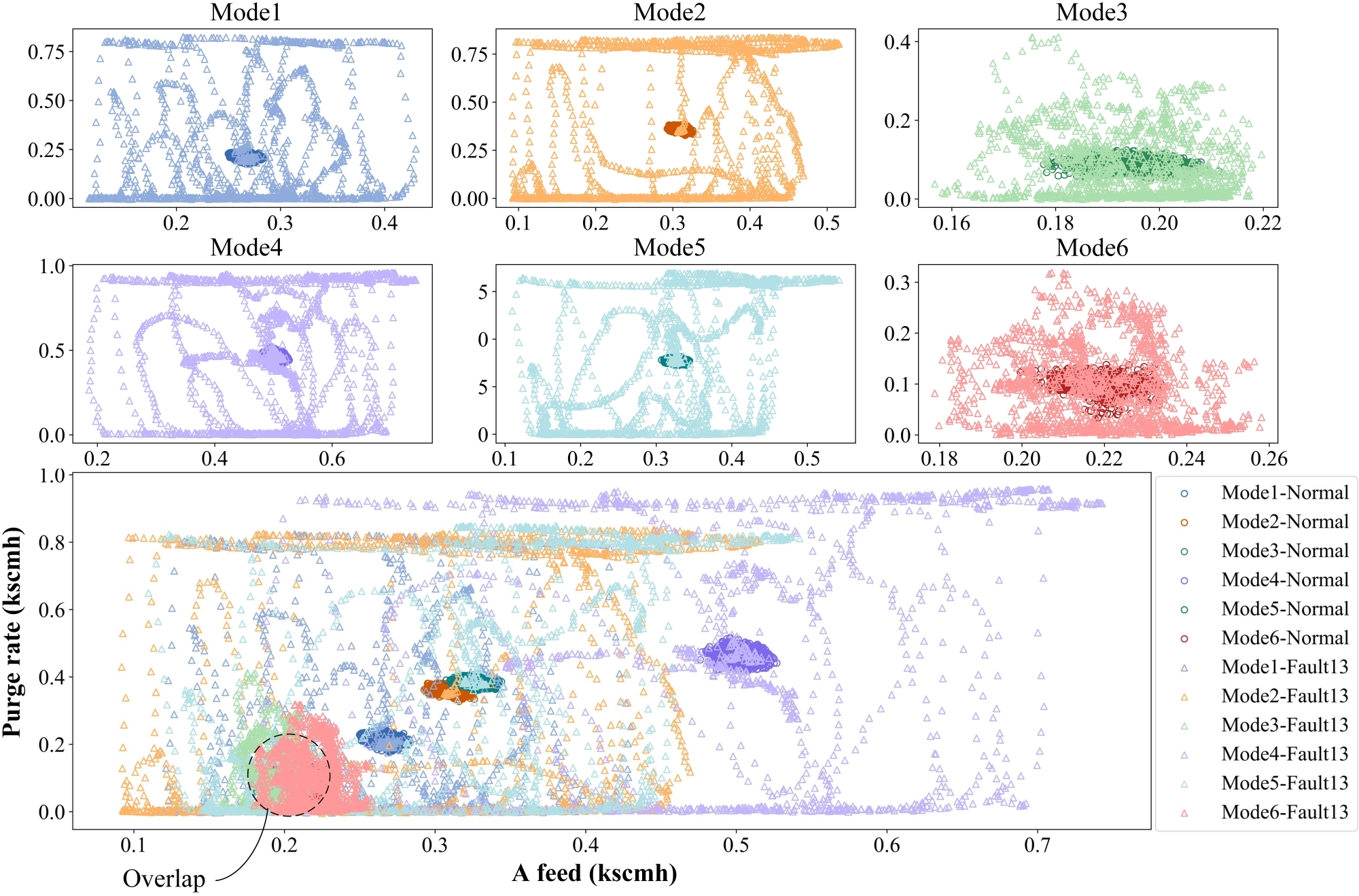}}
	\caption{Data distribution of two health conditions under six modes in the TE process.}
	\label{Fig1}
\end{figure}

In SMFD, both the training and test samples are collected from the same mode.
UMFD differs from SMFD in two main aspects. 
First, in UMFD, the operating mode of the sample is uncertain.
This means that although it is known that the sample belongs to one of the $M$ operating modes, the specific mode is not identifiable. 
In the case where the operating modes of the sample are known, a separate model can be built for each mode, and the corresponding model can be selected based on the sample's operating mode. 
Second, UMFD involves monitoring data from multiple modes in both the training and test samples.
Taking the two health conditions of the Tennessee Eastman (TE) process for demonstration, \hyperref[Fig1]{Fig. 1} presents the data distribution of two monitoring variables under six operating modes.
The circles and triangles denote the Normal condition and Fault 13, respectively. 
It is evident that, under different operating modes, samples belonging to the same health condition exhibit significant distributional differences. 
Additionally, the obvious overlap between the Normal condition and Fault 13 results in blurred category boundaries, thus making accurate classification difficult. 
The complex coupling between health conditions and operating modes in multimode process monitoring data puts higher requirements on the feature representation capability of the fault diagnosis model.

Cross-domain fault diagnosis focuses on the scenarios where there are distribution discrepancies between training and test samples.
It includes domain adaptation-based fault diagnosis (DAFD) and domain generalization-based fault diagnosis (DGFD).
The difference between the two is that DAFD relies on unlabeled test samples to train the model.
Compared to cross-domain fault diagnosis, UMFD does not involve domain shift between the training and test samples. 
However, the operating mode of the samples is inaccessible.
The extraction of universal features is key to cross-domain fault diagnosis. 
UMFD aims to develop models that can effectively learn shared representations reflecting system condition from monitoring data across multiple modes. 
Therefore, the research on UMFD plays an important role in cross domain fault diagnosis. 

\subsection{Depthwise convolution}
\begin{figure}[!ht]
	\centerline{\includegraphics[width=0.9\columnwidth,height=0.3\columnwidth]{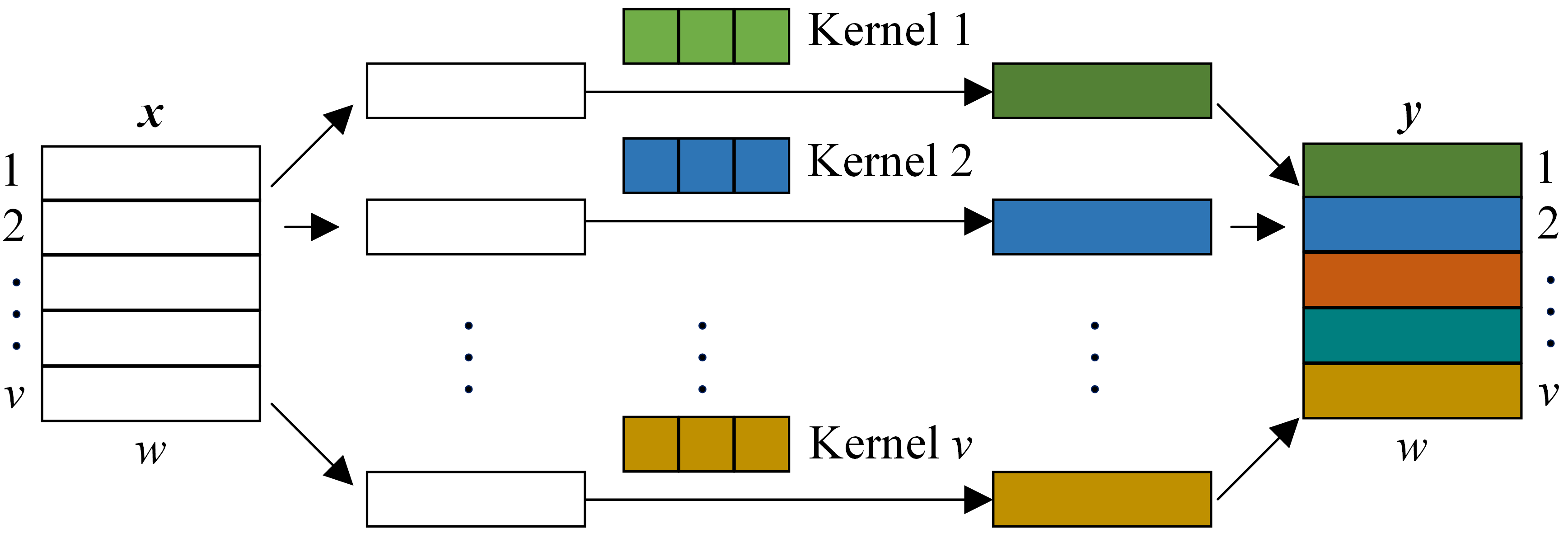}}
	\caption{Computation process of depthwise convolution.}
	\label{Fig2}
\end{figure}

In traditional convolution, a convolution kernel performs operations across all input channels to generate a single output channel. 
In contrast, depthwise convolution assigns a single convolution kernel to each input channel and processes each channel independently, which greatly reduces computational complexity \cite{RN260423}.
A depthwise convolution layer takes a feature map $\bm{x}  \in \mathbb{R}^{w\times v}$ as input and produces a feature map $\bm{y}  \in \mathbb{R}^{w\times v}$. $\bm{y}$ is formulated as $\bm{y}_{m,n}=\sum_i k_{m,i}\bm{x}_{m,n+i-1}$, where $k$ is the depthwise convolution kernel.
The computation process of depthwise convolution is illustrated in \hyperref[Fig2]{Fig. 2}.  

\subsection{GRU}
\begin{figure}[!ht]
	\centerline{\includegraphics[width=0.6\columnwidth,height=0.3122\columnwidth]{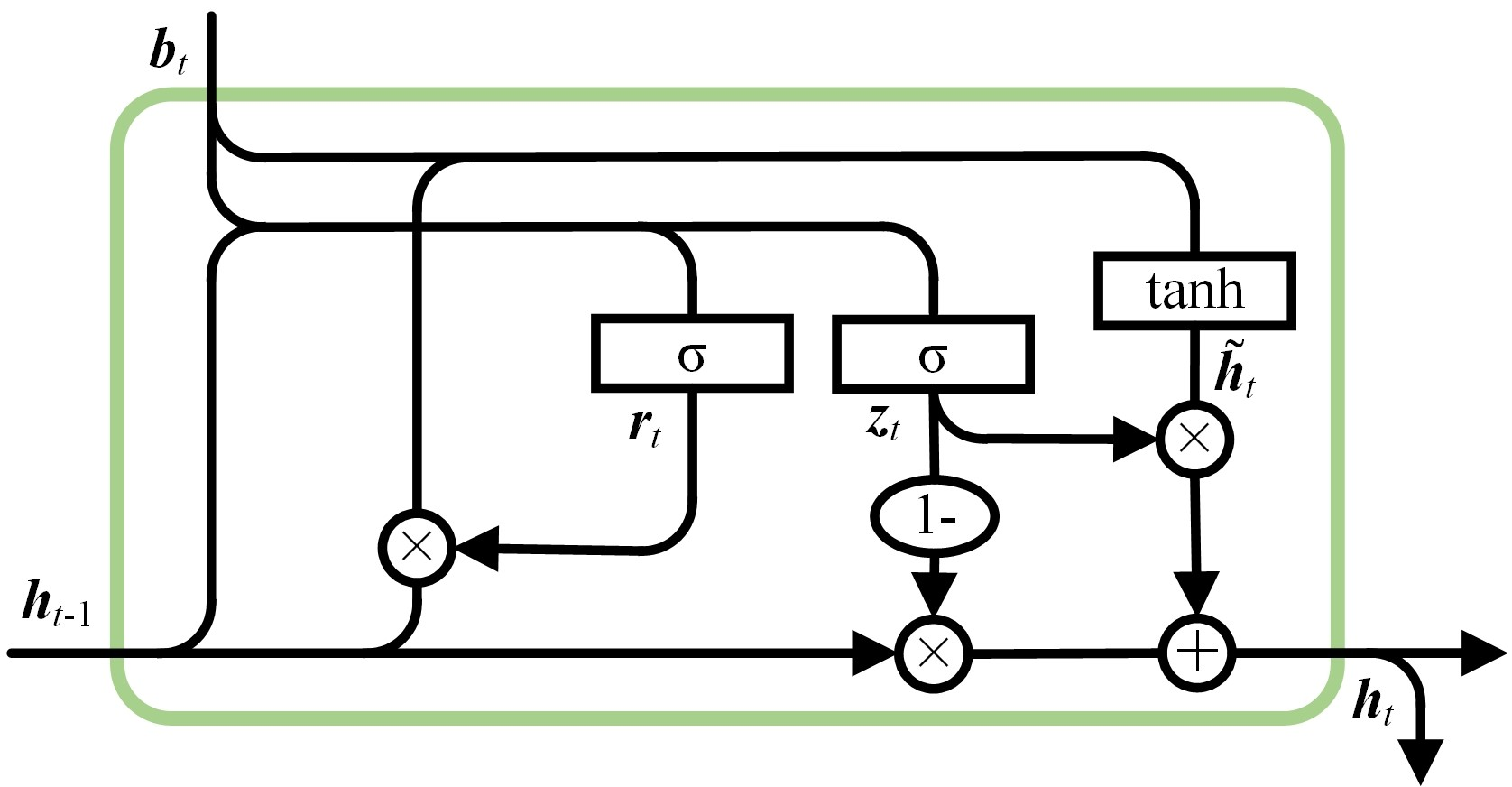}}
	\caption{Internal structure of GRU cell unit.}
	\label{Fig3}
\end{figure}

Compared to the CNN, the gated recurrent unit (GRU) have an advantage in processing time-series data.
The GRU has fewer parameters and achieve faster convergence \cite{RN260413}, and its structure is shown in \hyperref[Fig3]{Fig. 3}. 

The GRU takes the previous hidden state $\bm{h}_{t-1}$ and the current input $\bm{b}_{t}$ as inputs, and outputs the current hidden state $\bm{h}_{t}$.
The output $\bm{h}_{t}$ is formulated as follows \cite{RN260416},
\begin{equation}
	\begin{aligned}
		\bm{r}_{t} & = \sigma\left(\bm{W}_{r} \bm{b}_{t}+\bm{U}_{r} \bm{h}_{t-1}\right), \\
		\tilde{\bm{h}}_{t} & = \tanh \left(\bm{W} \bm{b}_{t}+\bm{U}\left(\bm{r}_{t} \odot \bm{h}_{t-1}\right)\right), \\
		\bm{z}_{t} & = \sigma\left(\bm{W}_{z} \bm{b}_{t}+\bm{U}_{z} \bm{h}_{t-1}\right), \\
		\bm{h}_{t} & = \left(1-\bm{z}_{t}\right) \odot \bm{h}_{t-1}+\bm{z}_{t} \odot \tilde{\bm{h}}_{t},
	\end{aligned}
	\label{Eq1}	
\end{equation}
where $\sigma$ denotes the Sigmoid activation function. 
$\bm{r}_{t}$ and $\bm{h}_{t}$ control the information reset and update processes, respectively. 
$\tilde{\bm{h}}_{t}$ represents the candidate hidden state. 
$\bm{W}_{r}$, $\bm{U}_{r}$, $\bm{W}$, $\bm{U}$, $\bm{W}_{z}$ and $\bm{U}_{z}$ are learnable weight matrices, which are jointly optimized across all time steps.

\section{Proposed method}
\label{Sec3}
\subsection{Overall structure of AMTFNet}
\label{subsec1}

\begin{figure}[!ht]
	\centerline{\includegraphics[width=1.0\columnwidth,height=0.9517\columnwidth]{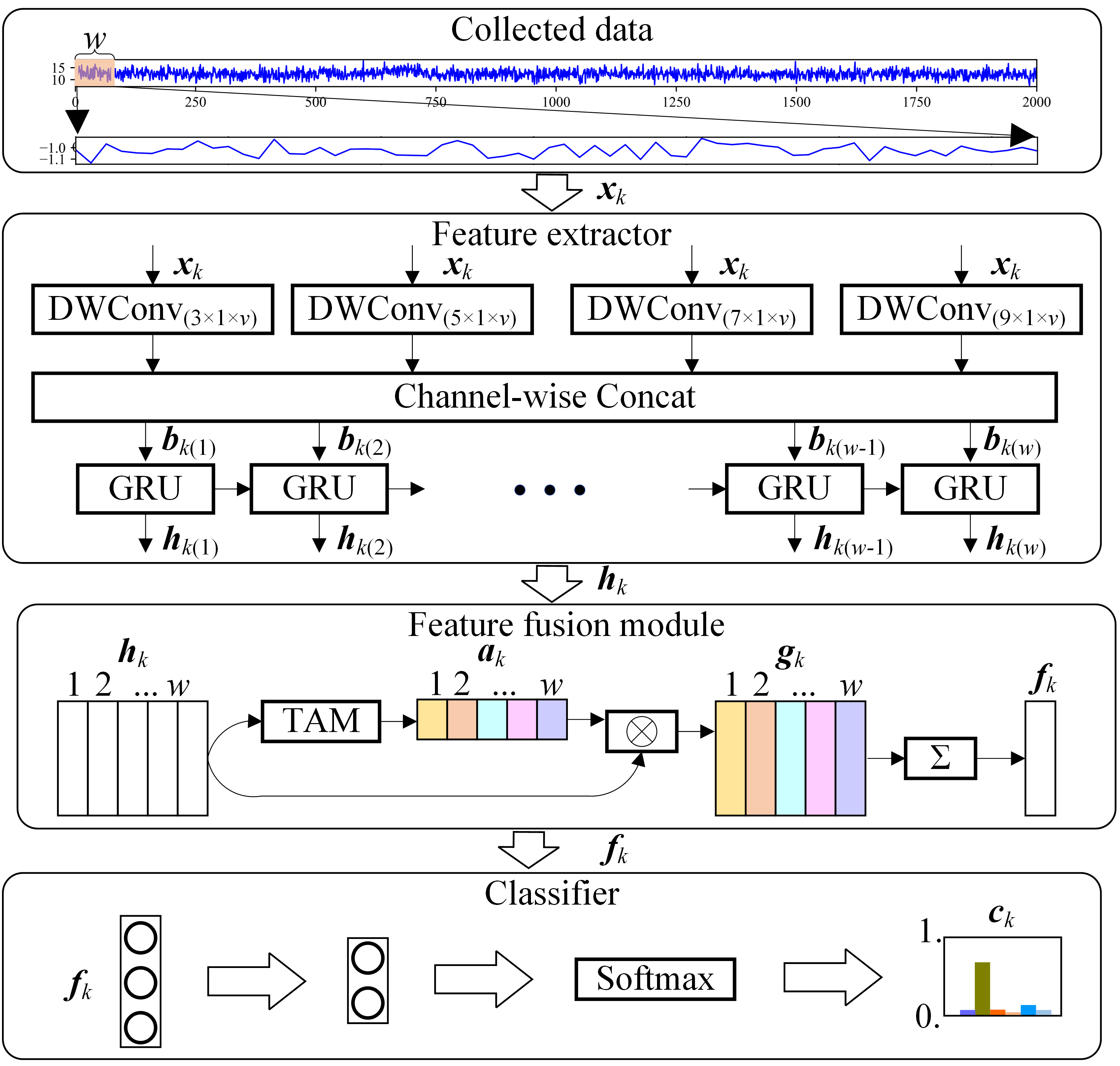}}
	\caption{Overall structure of AMTFNet.}
	\label{Fig4}
\end{figure}

The overall design of AMTFNet is shown in \hyperref[Fig4]{Fig. 4}. The model consists of three primary components: feature extractor $E$, feature fusion module $F$, and classifier $C$.  
$E$ is used to capture deep feature representations, after which $F$ performs weighted fusion of features across different time steps. 
Finally, $C$ is employed to determine the health condition of the system. 

\subsection{Feature extractor}
The feature extractor consists of MSDC, channel-wise concatenation and GRU.
MSDC is used to extract multiscale contextual local features in parallel using convolution kernels of different sizes.
Channel-wise concatenation is then employed to splice the multiscale local features along the channel dimension, followed by GRU to further extract deep long-term and short-term feature representations.

To preserve dynamic temporal information, the samples from the previous $w-1$ time steps are extended to the current time step $\bm{x}_k = [\bm{x}_{k-w+1}^\text{o},\bm{x}_{k-w+2}^\text{o},...,\bm{x}_{k}^\text{o}] \in \mathbb{R}^{v \times w}$.
To simplify the representation, let $\boldsymbol{x}_{k(i)}=\boldsymbol{x}_{k-w+i}^{\mathrm{o}}$, where $i=1,2, \ldots, w$. Then $\boldsymbol{x}_{k}=\left[\boldsymbol{x}_{k(1)}, \boldsymbol{x}_{k(2)}, \ldots, \boldsymbol{x}_{k(w)}\right]=\left[\boldsymbol{x}_{k-w+1}^{\mathrm{o}}, \boldsymbol{x}_{k-w+2}^{\mathrm{o}}, \ldots, \boldsymbol{x}_{k}^{\mathrm{o}}\right]$.
$\bm{x}_{k}$ is fed into MSDC and then passed through channel-wise concatenation. 
The output of $\bm{x}_{k}$ after the above processing is formulated as,
\begin{equation}
	\begin{aligned}
		\bm{b}_k = \text{Concat}[\text{DC}_{1\times n}(\bm{x}_k)], \quad n \in \{3, 5, 7, 9\},
	\end{aligned}
	\label{Eq2}	
\end{equation}
where $\bm{b}_k \in \mathbb{R}^{4v \times w}$, DC denotes the depthwise convolution operation, and the subscript indicates the kernel size.
In depthwise convolution, instance normalization is applied to weaken the mode features, and the activation function adopted is ReLU.
$\text{Concat}$ represents concatenation in the channel-wise direction. 

Then, $\bm{b}_k$ is fed into GRU, and the hidden vector at the $t$-th time step is formulated as follows,
\begin{equation}
	\begin{aligned}
		\bm{h}_{k(t)} = \text{GRU}(\bm{b}_{k(t)},\bm{h}_{k(t-1)}).
	\end{aligned}
	\label{Eq3}	
\end{equation}

The extracted features corresponding to the $k$-th sample is formulated as, $\bm{h}_k=E\left(\bm{x}_k\right)=[\bm{h}_{k(1)},\bm{h}_{k(2)},...,\bm{h}_{k(w)}]$.

\subsection{Feature fusion module}
Considering that the amount of domain-invariant information varies across different time points, a temporal attention mechanism (TAM) is incorporated into the feature fusion module to help the model concentrate on features at key time points. 
The feature fusion module takes the feature map $\bm{h}_k$ as input and infers a temporal attention map $\bm{a}_k$ through the TAM, as shown in \hyperref[Fig4]{Fig. 4}. $\bm{a}_k$ is then used as a weight representing the importance of each time step to obtain the feature fusion result $\bm{f}_k$. The overall process of feature fusion can be described as follows,
\begin{equation}
	\begin{aligned}
		\bm{f}_{k}= F (\bm{h}_{k})
		=\sum_{t=1}^{w} \bm{g}_{k(t)}=\sum_{t=1}^{w} \bm{h}_{k(t)} \otimes \bm{a}_{k(t)},
	\end{aligned}
	\label{Eq4}	
\end{equation}
where $\otimes$ denotes element-wise multiplication. 

\begin{figure}[!ht]
	\centerline{\includegraphics[width=1.\columnwidth,height=0.2632\columnwidth]{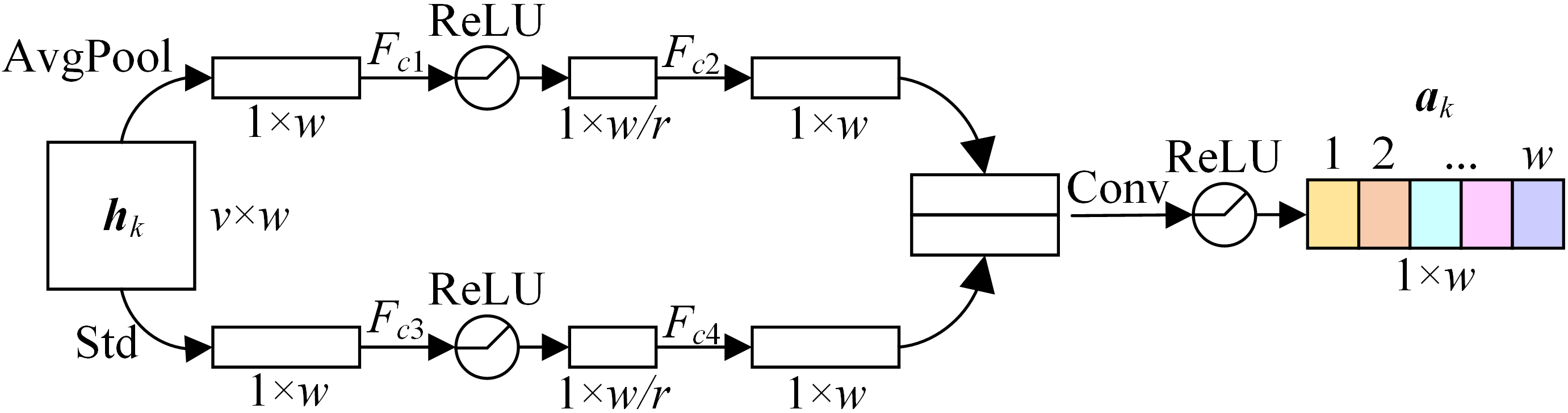}}
	\caption{Structure of TAM.}
	\label{Fig5}
\end{figure}

The structure of the TAM is shown in \hyperref[Fig5]{Fig. 5}.
The relationships between different time steps of the features are utilized to generate the temporal attention map.
Average pooling and standard deviation are used to aggregate information along the variable dimension of the feature map.
These aggregated features are then transmitted to their respective fully connected layers to generate the average-pooling attention map $\bm{p}_1$ and the standard deviation attention map $\bm{p}_2$.
The calculations for $\bm{p}_1$ and $\bm{p}_2$ are formulated as,
\begin{equation}
	\begin{aligned}
		\bm{p}_1&=F_{c2} \left( \sigma_1 \left( F_{c1}(\text{AvgPool}\left(\bm{h}_k\right))\right)\right), \\
        \bm{p}_2&=F_{c4} \left( \sigma_1 \left( F_{c3}(\text{Std}\left(\bm{h}_k\right))\right)\right), 
	\end{aligned}
	\label{Eq5}	
\end{equation}
where $\sigma_1$ denotes the ReLU function, AvgPool denote the average-pooling operations, Std denotes the calculation of standard deviation, $F_{c1}-F_{c4}$ denote the fully connected layers.

The average-pooling and standard deviation attention maps describe the attention to the feature map at different time steps from the perspectives of mean and variability, respectively. 
A convolution neural network is employed to fuse these two dimensions of attention and generate the final temporal attention. 
The temporal attention map $\bm{a}_k$ is formulated as,
\begin{equation}
	\begin{aligned}
		\bm{a}_k=\sigma_2\left(\text{Conv}\left( \text{Concat}\left(\bm{p}_1;\bm{p}_2\right)\right)\right),
	\end{aligned}
	\label{Eq6}	
\end{equation}
where Conv denotes the convolution operation, $\sigma_2$ denotes ReLU function. 
\subsection{Model output and application process}
The classifier takes the fused feature $\bm{f}_k$ as input and infers health condition class.
The model output $\bm{c}_{k}$ is obtained via a fully connected layer, and is formulated as,
\begin{equation}
	\begin{aligned}
		\bm{o}_{k} &= F_{c}\left(\bm{f}_{k}\right) = \left[o_{k, 1}, o_{k, 2}, \ldots, o_{k, L}\right], \\
		c_{k, l} &= C\left(f_{k, l}\right) = \operatorname{Softmax}\left(o_{k, l}\right) = \frac{\exp \left(o_{k, l}\right)}{\sum_{j = 1}^{L} \exp \left(o_{k, j}\right)},
	\end{aligned}
	\label{Eq7}	
\end{equation}
where $l=1,2,...,L$, $F_{c}$ denotes the fully connected layer. $c_{k, l}$ represents the $l$-th output component of the output for the $k$-th sample, which indicates the predicted confidence that the sample is assigned to category $l$. 
In addition, dropout is added to prevent overfitting.
The cross-entropy loss is applied as the objective function for the fault diagnosis model, and it is defined as,

\begin{equation}
	\begin{aligned}
		L=-\sum_{k=1}^{N_b}\sum_{l=1}^{L}y_{k,l}\log(c_{k,l}),
	\end{aligned}
	\label{Eq8}	
\end{equation}
where $N_b$ denotes the batch size, and $y_{k,l}$ takes the value 0 or 1, which indicates whether the label of the $k$-th sample is $l$. 

\begin{figure}[!ht]
	\centerline{\includegraphics[width=1\columnwidth,height=0.728\columnwidth]{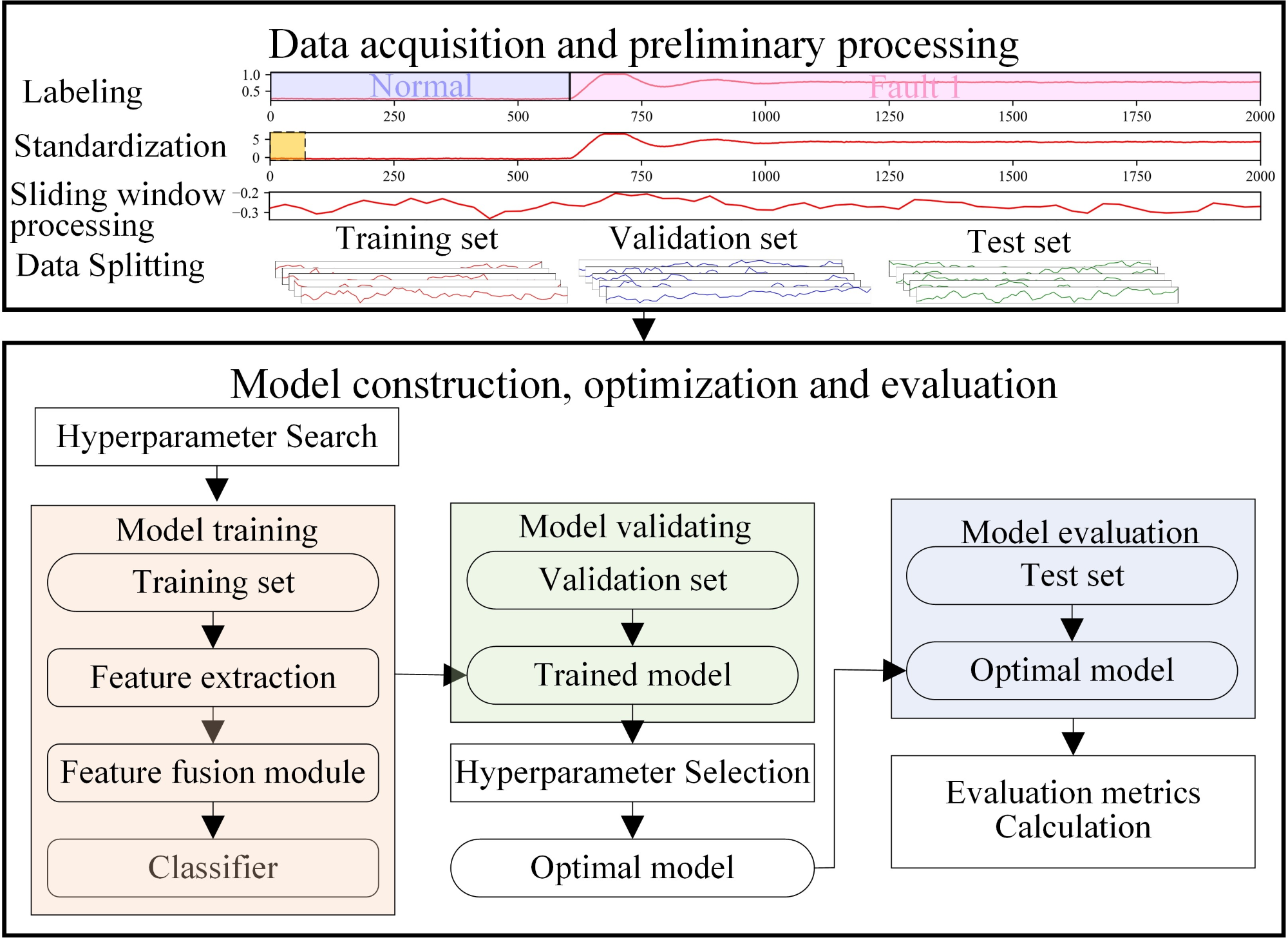}}
	\caption{The process of applying AMTFNet model in uncertain-mode fault diagnosis.}
	\label{Fig6}
\end{figure}

The process of applying the AMTFNet model to fault diagnosis is illustrated in  \hyperref[Fig6]{Fig. 6}, which mainly includes two steps: the first is data acquisition and preliminary processing, and the second is model construction, optimization, and evaluation. 
The detailed description is provided below.

(1) Data acquisition and preliminary processing mainly includes data acquisition, labeling, standardization, sliding window processing, and data splitting. 
Firstly, the monitoring data across multiple modes are collected, and each sample is assigned a corresponding category label.
Then, z-score normalization is applied to standardize samples across all modes.
The standardized result is formulated as $(\bm{x}_{k}^\text{o} - \bm{\mu})/\bm{\sigma}$, where $\bm{\mu}$ and $\bm{\sigma}$ denote the mean and standard deviation estimated from the fault-free samples across all modes. 
And sliding window of size 64 is employed to preserve dynamic temporal information.
After applying the sliding window technique, only the label of the last time step within each window is treated as the target to be predicted at the current time point.
The stratified sampling strategy was employed to split the dataset according to the health condition categories in each operating mode. Specifically, 80\% of the samples in each category were assigned to the training set, while the remaining 20\% were equally divided into the validation and test sets.
Since the operating mode associated with each sample is uncertain, this information cannot be utilized during the model training phase.

(2) 
Model construction and optimization involves training the fault diagnosis model and selecting the best-performing model. 
Model evaluation is performed on the test set to measure the overall effectiveness of the fault diagnosis model.

\section{Experiments}
\label{Sec4}
\subsection{Evaluation metrics}
Micro-F1, Macro-F1, Fault Diagnosis Rate (FDR), and False Positive Rate (FPR) are selected to compare the model’s ability in diagnosing faults.
To facilitate the presentation of the calculation formula for each evaluation metric, the confusion matrix corresponding to the $l$-th category is provided in \hyperref[Table2]{Table 2}, where each element denotes the quantity of data instances matching the relevant condition.

\begin{table}[!ht]
	\centering
    \small
	\begin{threeparttable}
	\caption{Confusion matrix corresponding to the $l$-th category.}
	\label{Table2}
	\begin{tabular}{lll}
		\hline
		\textbf{}                                                                                           & \textbf{\begin{tabular}[c]{@{}l@{}}Inferred category is $l$\end{tabular}} & \textbf{\begin{tabular}[c]{@{}l@{}}Inferred category is not $l$\end{tabular}} \\ \hline
		\textbf{Real category is $l$}    & $TP_l$                                                                                             & $FN_l$                                                                                                 \\
		\textbf{Real category is not $l$} & $FP_l$                                                                                             & $TN_l$                                                                                               \\ \hline
	\end{tabular}

\end{threeparttable}
    
\end{table}

The Micro-F1 is computed as follows \cite{RN260427},
\begin{equation}
	\begin{aligned}
		&\text { Precision }_{\text {Micro }} = \frac{\sum_{l = 1}^{L} \mathrm{TP}_{l}}{\sum_{l = 1}^{L} \mathrm{TP}_{l}+\sum_{l = 1}^{L} \mathrm{FP}_{l}}, \\
		&\text { Recall }_{\text {Micro }} = \frac{\sum_{l = 1}^{L} \mathrm{TP}_{l}}{\sum_{l = 1}^{L} \mathrm{TP}_{l}+\sum_{l = 1}^{L} \mathrm{FN}_{l}},\\
	    &\text { F1 }_{\text {Micro }} = 2 \cdot \frac{\text { Precision }_{\text {Micro }} \cdot \text { Recall }_{\text {Micro }}}{\text { Precision }_{\text {Micro }}+\text { Recall }_{\text {Micro }}}. 
    \end{aligned}	
	\label{Eq9}	
\end{equation}
Micro-F1 is calculated at the dataset level and each sample has the equal weight.

The Macro-F1 is computed as follows \cite{RN260427},
\begin{equation}
	\begin{aligned}
		&\text { Precision }_{l} = \frac{\mathrm{TP}_{l}}{ \mathrm{TP}_{l}+ \mathrm{FP}_{l}}, \\
		&\text { Recall }_{l} = \frac{ \mathrm{TP}_{l}}{ \mathrm{TP}_{l}+ \mathrm{FN}_{l}},\\
		&\text { F1 }_{l} = 2 \cdot \frac{\text { Precision }_{l} \cdot \text { Recall }_{l}}{\text { Precision }_{l}+\text { Recall }_{l}}, \\
		&\text { F1 }_{\text {macro }} = \frac{\sum_{l = 1}^{L} \text { F1 }_{l}}{L}. \\
	\end{aligned}	
	\label{Eq10}	
\end{equation}
Macro-F1 is calculated at the class level and each class has the equal weight.

The FDR and FPR are computed as follows \cite{RN260387},
\begin{equation}
	\begin{aligned}
		\text{FDR}_l &= \frac{ \mathrm{TP}_{l}}{ \mathrm{TP}_{l}+ \mathrm{FN}_{l}},\\
		\text{FPR}_l &= \frac{ \mathrm{FP}_{l}}{ \mathrm{FP}_{l}+ \mathrm{TN}_{l}}.\\
	\end{aligned}	
	\label{Eq11}	
\end{equation}
FDR measures the ratio of instances with the real category $l$ that are correctly classified as $l$. FPR measures the ratio of instances from other categories are mistakenly classified as category $l$.

Additionally, both model size and runtime are considered as metrics to examine the scalability of the fault diagnosis methods.

\subsection{Implementation details}
The detailed structure of the modules in AMTFNet is shown in \hyperref[Table3]{Table 3}. 
The model is trained for 30 epochs using a batch size of 512. The initial learning rate is set to 0.01 and is exponentially decayed by a factor of 0.3 every three epochs to facilitate convergence.
\begin{table}[!ht]
	\centering
	\small
	\caption{Structure of the modules in AMTFNet.}
	\label{Table3}
	\begin{tabular}{llll}
		\hline
		\textbf{Modules}        & \textbf{Symbols}       & \textbf{Type}   & \textbf{Input, output, kernel size} \\ \hline
		\textbf{Feature}        & $\text{DC}_{1\times3}$ & Depthwise conv  & ($v$$\times$64,$v$$\times$64,(3$\times$1)$\times$$v$)           \\
		\textbf{extractor $E$}  & $\text{DC}_{1\times5}$ & Depthwise conv  & ($v$$\times$64,$v$$\times$64,(5$\times$1)$\times$$v$)           \\
		& $\text{DC}_{1\times7}$ & Depthwise conv  & ($v$$\times$64,$v$$\times$64,(7$\times$1)$\times$$v$)           \\
		& $\text{DC}_{1\times9}$ & Depthwise conv  & ($v$$\times$64,$v$$\times$64,(9$\times$1)$\times$$v$)           \\
		& -                      & GRU             & ($4v$$\times$64, 100$\times$64, -)                \\
		\textbf{Feature}        & $F_{c1}$               & Fully connected & (64, 64/$r$, -)                     \\
		\textbf{fusion $F$}     & $F_{c2}$               & Fully connected & (64/$r$, 64, -)                     \\
		& Conv                   & Conv1d          & (3$\times$64, 1$\times$64, 3)                     \\
		\textbf{Classifier $C$} & $F_{c3}$               & Fully connected & (100,$c$, -)                        \\ \hline
	\end{tabular}
\end{table}

To demonstrate the capability of the proposed AMTFNet, several advanced models are applied for comparison. The models used for comparison are as follows:


(1) CNN-LSTM \cite{RN260391}. 
CNN and LSTM are fused in this model to collaboratively capture features indicative of health conditions.

(2) IPO-ViT \cite{RN260425}. 
This model employed a Transformer-based architecture that 
utilized the self-attention mechanism to capture global features.

(3) MCNN-DBiGRU \cite{RN260405}.  This model integrates the multi-scale convolution neural network with the deep bidirectional GRU to  extract discriminative feature representations.

(4) MGAMN \cite{RN260403}. 
 This model was constructed by combining data augmentation and representation learning for fault diagnosis under domain shift conditions.

Methods 1-3 have demonstrated superior performance in SMFD, while Method 4 represents a state-of-the-art DGFD. 
Advanced SMFD and DGFD methods are employed in UMFD scenario to validate the challenges in such settings.
The comprehensive comparison of all methods is conducted to validate the effectiveness of the developed model.

The model setup details for the ablation experiments are shown in \hyperref[Table4]{Table 4}.
In the feature extractor, models A1 and A3 employ MSDC, while models A2 and A4 utilize GRU. 
Models A5 and A6 integrates both MSDC and GRU for feature learning. 
In the feature fusion module, different attention mechanisms are employed to validate the effectiveness of the proposed TAM. 
Specifically, TAM is integrated into models A3 and A4, while the Squeeze-and-Excitation (SE) block \cite{RN266425} is incorporated into model A6 to enhance the feature representation.
All models perform classification through fully connected layers.

\begin{table}[!ht]
	\centering
	\caption{Model setup details for the ablation experiments.}
	\label{Table4}
\begin{tabular}{llll}
\hline
\multirow{2}{*}{\textbf{Model}} & \multicolumn{2}{l}{\textbf{Feature extractor}} & \multirow{2}{*}{\textbf{Feature fusion module}} \\ \cline{2-3}
                                & \textbf{MSDC}          & \textbf{GRU}          &                                                 \\ \hline
\textbf{A1}                     & $\checkmark$           & -                     & -                                               \\
\textbf{A2}                     & -                      & $\checkmark$          & -                                               \\
\textbf{A3}                     & $\checkmark$           & -                     & TAM                                             \\
\textbf{A4}                     & -                      & $\checkmark$          & TAM                                             \\
\textbf{A5}                     & $\checkmark$           & $\checkmark$          & -                                               \\
\textbf{A6}                     & $\checkmark$           & $\checkmark$          & SE block                                        \\ \hline
\end{tabular}
\end{table}

\subsection{Case one: TE process}

\subsubsection{Task description}
\begin{figure}[!ht]
	\centering
\includegraphics[width=1.\textwidth,height=0.5632\textwidth]{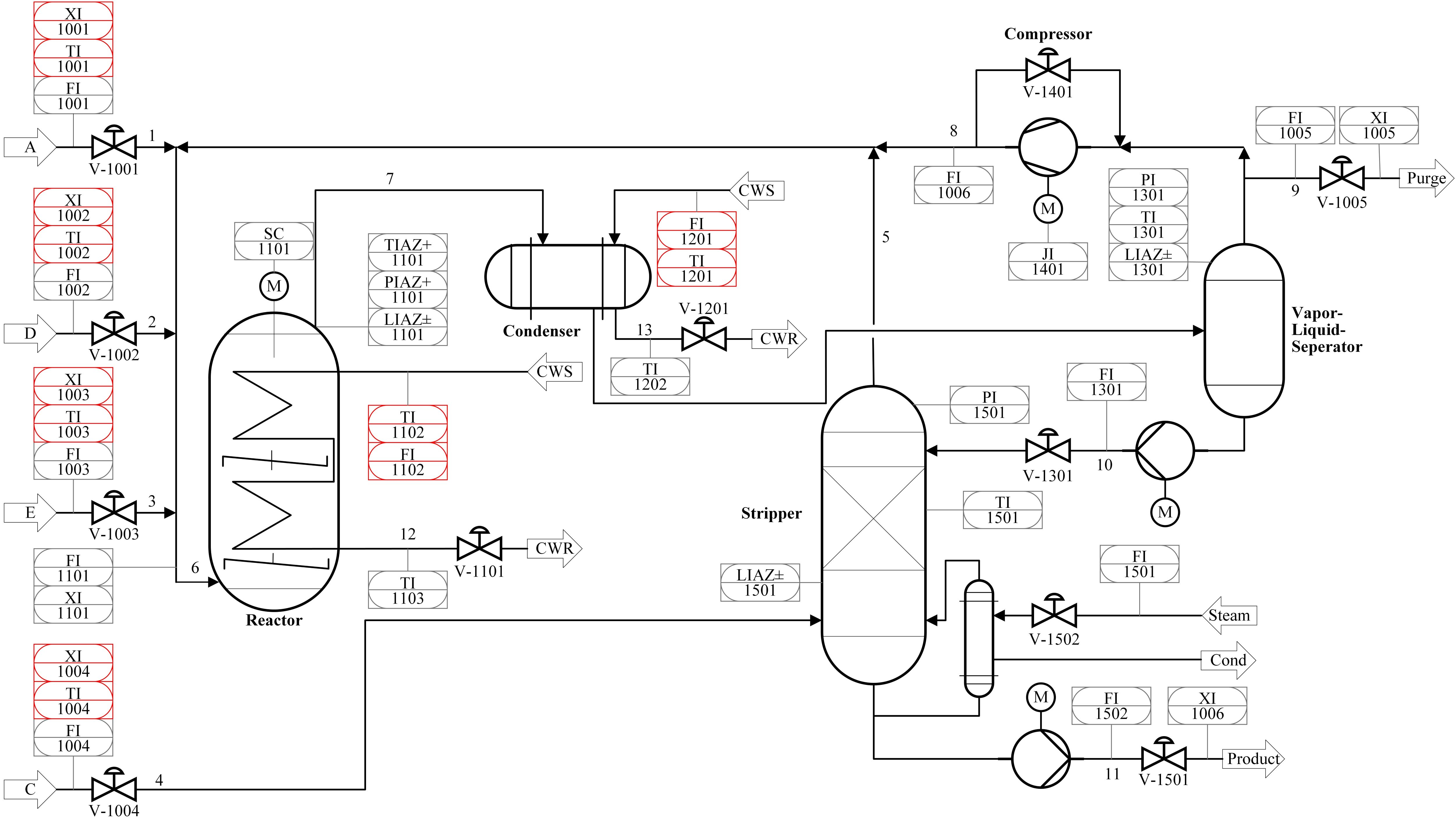}
	\caption{P\&ID of the revised process model \cite{RN260408}.}
	\label{Fig7}
\end{figure}

The TE process simulation model has become a benchmark for industrial system fault diagnosis \cite{RN260406}. 
Bathelt et al. \cite{RN260408} introduced additional process measurements and disturbances to the TE process, as shown in \hyperref[Fig7]{Fig. 7}. 
This process includes 41 measured variables and 12 manipulated variables.
It is capable of simulating 28 faults under 6 modes. 
Liu et al. obtained the multimode fault diagnosis datasets by adjusting the parameters of the TE process simulation models \cite{RN260410}.
Since Fault 3 and Fault 16 are analogous with the normal operating conditions \cite{RN260409,RN266423}, they are not included in this study.
The system health condition categories, modes and task settings are presented in \hyperref[Table5]{Table 5} and \hyperref[Table6]{Table 6}. In tasks T1 to T6, each task's dataset consists of data from five modes, with the corresponding positions marked as $\checkmark$.

\begin{table}[!ht]
	\centering
	\caption{Faults of TE process used in uncertain-mode fault diagnosis \cite{RN260407,RN260408}.}
	\label{Table5}
    \fontsize{10}{14}\selectfont
	\begin{tabular}{lll}
\hline
\textbf{No.} & \textbf{Description}                                     & \textbf{Type}    \\ \hline
N            & Normal                                                   & -                \\
F1           & A/C feed ratio, B composition constant (stream 4)        & Step             \\
F2           & B composition, A/C ratio constant (Stream 4)             & Step             \\
F4           & Reactor cooling water inlet temperature                  & Step             \\
F5           & Condenser cooling water inlet temperature                & Step             \\
F6           & A feed loss (stream 1)                                   & Step             \\
F7           & C header pressure loss - reduced availability (stream 4) & Step             \\
F8           & A, B, C feed composition (stream 4)                      & Random variation \\
F9           & D feed temperature (stream 2)                            & Random variation \\
F10          & C feed temperature (stream 4)                            & Random variation \\
F11          & Reactor cooling water inlet temperature                  & Random variation \\
F12          & Condenser cooling water inlet temperature                & Random variation \\
F13          & Reaction kinetics                                        & Drift            \\
F14          & Reactor cooling water valve                              & Sticking         \\
F15          & Condenser cooling water valve                            & Sticking         \\
F17-20       & Unknown                                                  & Unknown          \\
F21          & A feed temperature (stream1)                             & Random variation \\ \hline
\end{tabular}
\end{table}

\begin{table}[!ht]
	\centering
	\caption{Modes of TE process used in uncertain-mode fault diagnosis \cite{RN260407}.}
	\label{Table6}

\begin{tabular}{llllllll}
\hline
No. & G/H mass ratio and production rate & T1           & T2           & T3           & T4           & T5           & T6           \\ \hline
M1       & 50/50, G: 7038kg/h, H: 7038kg/h    &              & $\checkmark$ & $\checkmark$ & $\checkmark$ & $\checkmark$ & $\checkmark$ \\
M2       & 10/90, G: 1408kg/h, H: 12669kg/h   & $\checkmark$ &              & $\checkmark$ & $\checkmark$ & $\checkmark$ & $\checkmark$ \\
M3       & 90/10, G: 10000kg/h, H: 1111kg/h   & $\checkmark$ & $\checkmark$ &              & $\checkmark$ & $\checkmark$ & $\checkmark$ \\
M4       & 50/50, maximum production rate     & $\checkmark$ & $\checkmark$ & $\checkmark$ &              & $\checkmark$ & $\checkmark$ \\
M5       & 10/90, maximum production rate     & $\checkmark$ & $\checkmark$ & $\checkmark$ & $\checkmark$ &              & $\checkmark$ \\
M6       & 90/10, maximum production rate     & $\checkmark$ & $\checkmark$ & $\checkmark$ & $\checkmark$ & $\checkmark$ &              \\ \hline
\end{tabular}
\end{table}

The TE dataset used in the experiment was generated through SIMULINK simulation of the TE process, covering 19 system health state categories across 6 modes. 
The simulation was run for 100 hours with samples collected every 3 minutes and the fault was introduced at the 30th hour.
Note that during fault simulations in some modes, the simulation process might stop abruptly, resulting in fewer samples for some categories, leading to class imbalance. 
Specifically, such abrupt terminations occurred during the simulations of F6 in M1, M2, M3, M5, and M6, as well as F1, F6, and F7 in M4.

\subsubsection{Experiment results}
The experiment results for Micro F1 and Macro F1 scores are presented in \hyperref[Table7]{Table 7} and \hyperref[Table8]{Table 8}, respectively.
AMTFNet achieves the highest average Micro F1 and average Macro F1 scores among the five fault diagnosis models, with values of 0.9792 and 0.9803, respectively.
Compared to the CNN-LSTM, IPO-ViT, MCNN-DBiGRU and MGAMN models, AMTFNet achieves a 3.18\%, 9.08\%, 7.10\% and 15.47\% increase in average Micro F1 scores, respectively, and a 2.97\%, 8.48\%, 6.47\% and 14.19\% improvement in average Macro F1 scores, respectively.
Taking Task T1 as an illustration, the experiment results for FDR and FPR scores are shown in \hyperref[Table9]{Table 9}.
Although previous studies have found that Fault 9 and Fault 15 are difficult to diagnose \cite{RN260409,RN266423}, the proposed model has FDR values of more than 0.98 for both fault types.
This demonstrates the model's superior ability in  capturing highly discriminative fault features.
AMTFNet model achieved the highest FDR, with FDR values exceeding 0.9 for both the Normal and Fault 21 categories, whereas the comparative models showed FDR values below 0.8 for these categories.
In addition, the AMTFNet model achieved the lowest FPR. 
This validates that the proposed AMTFNet model can effectively capture deep features in multimode distributed data, enabling efficient fault diagnosis under uncertain modes.

\begin{table}[!ht]
	\centering
    \begin{threeparttable}
	\caption{Micro F1 scores of different methods on the TE process dataset.}
	\label{Table7}
	\begin{tabular}{cccccc}
\hline
\multicolumn{1}{l}{\multirow{2}{*}{\textbf{Task}}} & \multicolumn{5}{c}{\textbf{Model}}                                                              \\ \cline{2-6} 
\multicolumn{1}{l}{}                               & \textbf{CNN-LSTM} & \textbf{IPO-ViT} & \textbf{MCNN-DBiGRU} & \textbf{MGAMN} & \textbf{AMTFNet} \\ \hline
\textbf{T1}                                        & 0.9595            & 0.8941           & 0.9152               & 0.8718         & \textbf{0.9824}  \\
\textbf{T2}                                        & 0.9236            & 0.9015           & 0.9375               & 0.8501         & \textbf{0.9758}  \\
\textbf{T3}                                        & 0.9526            & 0.8939           & 0.8905               & 0.8376         & \textbf{0.9789}  \\
\textbf{T4}                                        & 0.9709            & 0.9176           & 0.9118               & 0.8602         & \textbf{0.9814}  \\
\textbf{T5}                                        & 0.9488            & 0.8989           & 0.9031               & 0.8326         & \textbf{0.9777}  \\
\textbf{T6}                                        & 0.9385            & 0.8802           & 0.9277               & 0.8356         & \textbf{0.9789}  \\
\textbf{Avg}                                       & 0.9490            & 0.8977           & 0.9143               & 0.8480         & \textbf{0.9792}  \\ \hline
\end{tabular}

\end{threeparttable}
\end{table}

\begin{table}[!ht]
	\centering
    \begin{threeparttable}
	\caption{Macro F1 scores of different methods on the TE process dataset.}
	\label{Table8}
	\begin{tabular}{cccccc}
\hline
\multicolumn{1}{l}{\multirow{2}{*}{\textbf{Task}}} & \multicolumn{5}{c}{\textbf{Model}}                                                              \\ \cline{2-6} 
\multicolumn{1}{l}{}                               & \textbf{CNN-LSTM} & \textbf{IPO-ViT} & \textbf{MCNN-DBiGRU} & \textbf{MGAMN} & \textbf{AMTFNet} \\ \hline
\textbf{T1}                                        & 0.9622            & 0.9008           & 0.9211               & 0.8809         & \textbf{0.9835}  \\
\textbf{T2}                                        & 0.9280            & 0.9077           & 0.9416               & 0.8621         & \textbf{0.9774}  \\
\textbf{T3}                                        & 0.9550            & 0.8992           & 0.9009               & 0.8503         & \textbf{0.9795}  \\
\textbf{T4}                                        & 0.9723            & 0.9216           & 0.9171               & 0.8663         & \textbf{0.9823}  \\
\textbf{T5}                                        & 0.9518            & 0.9055           & 0.9110               & 0.8422         & \textbf{0.9791}  \\
\textbf{T6}                                        & 0.9424            & 0.8876           & 0.9327               & 0.8492         & \textbf{0.9802}  \\
\textbf{Avg}                                       & 0.9520            & 0.9037           & 0.9207               & 0.8585         & \textbf{0.9803}  \\ \hline
\end{tabular}

\end{threeparttable}
\end{table}

\begin{table}[!ht]
	\centering
	\caption{FDR and FPR scores of different methods on the TE process dataset.}
    
	\label{Table9}
    \begin{adjustbox}{width=1.4\textwidth,center=\textwidth}
\begin{tabular}{lllllllllll}
\hline
\textbf{}      & \multicolumn{2}{c}{\textbf{CNN-LSTM}} & \multicolumn{2}{c}{\textbf{IPO-ViT}} & \multicolumn{2}{c}{\textbf{MCNN-DBiGRU}} & \multicolumn{2}{c}{\textbf{MGAMN}} & \multicolumn{2}{c}{\textbf{AMTFNet}} \\
\textbf{Class} & \textbf{FDR}      & \textbf{FPR}      & \textbf{FDR}      & \textbf{FPR}     & \textbf{FDR}        & \textbf{FPR}       & \textbf{FDR}     & \textbf{FPR}    & \textbf{FDR}      & \textbf{FPR}     \\ \hline
\textbf{N}     & 0.7686            & 0.0136            & 0.5443            & 0.0281           & 0.4700              & 0.0316             & 0.3714           & 0.0305          & 0.9186            & 0.0061           \\
\textbf{F1}    & 0.9983            & 0.0002            & 0.9948            & 0.0009           & 0.9948              & 0.0000             & 0.9930           & 0.0004          & 1.0000            & 0.0002           \\
\textbf{F2}    & 0.9986            & 0.0000            & 0.9914            & 0.0004           & 0.9986              & 0.0001             & 0.9900           & 0.0003          & 0.9986            & 0.0004           \\
\textbf{F4}    & 0.9986            & 0.0000            & 0.9957            & 0.0000           & 1.0000              & 0.0001             & 0.9986           & 0.0000          & 1.0000            & 0.0000           \\
\textbf{F5}    & 0.9957            & 0.0000            & 0.9929            & 0.0003           & 0.9957              & 0.0002             & 0.9929           & 0.0003          & 0.9971            & 0.0000           \\
\textbf{F6}    & 1.0000            & 0.0000            & 1.0000            & 0.0000           & 1.0000              & 0.0001             & 0.9836           & 0.0001          & 1.0000            & 0.0000           \\
\textbf{F7}    & 1.0000            & 0.0000            & 1.0000            & 0.0000           & 1.0000              & 0.0000             & 1.0000           & 0.0001          & 1.0000            & 0.0000           \\
\textbf{F8}    & 0.9914            & 0.0004            & 0.9671            & 0.0007           & 0.9886              & 0.0001             & 0.9786           & 0.0004          & 0.9929            & 0.0003           \\
\textbf{F9}    & 0.9300            & 0.0054            & 0.7429            & 0.0176           & 0.8071              & 0.0033             & 0.7514           & 0.0311          & 0.9829            & 0.0013           \\
\textbf{F10}   & 0.9800            & 0.0019            & 0.8929            & 0.0085           & 0.9786              & 0.0030             & 0.9214           & 0.0023          & 0.9886            & 0.0008           \\
\textbf{F11}   & 0.9857            & 0.0002            & 0.9743            & 0.0002           & 0.9871              & 0.0002             & 0.9843           & 0.0000          & 0.9957            & 0.0002           \\
\textbf{F12}   & 0.9957            & 0.0003            & 0.9900            & 0.0004           & 0.9957              & 0.0000             & 0.9843           & 0.0002          & 0.9971            & 0.0002           \\
\textbf{F13}   & 0.9814            & 0.0008            & 0.9686            & 0.0020           & 0.9743              & 0.0002             & 0.9529           & 0.0006          & 0.9857            & 0.0013           \\
\textbf{F14}   & 0.9929            & 0.0003            & 0.9800            & 0.0019           & 0.9543              & 0.0046             & 0.9929           & 0.0001          & 0.9943            & 0.0000           \\
\textbf{F15}   & 0.9914            & 0.0010            & 0.6943            & 0.0187           & 0.8614              & 0.0089             & 0.5714           & 0.0387          & 1.0000            & 0.0001           \\
\textbf{F17}   & 0.9757            & 0.0008            & 0.9771            & 0.0010           & 0.9757              & 0.0003             & 0.9700           & 0.0002          & 0.9771            & 0.0002           \\
\textbf{F18}   & 0.9500            & 0.0019            & 0.9357            & 0.0010           & 0.9443              & 0.0034             & 0.9357           & 0.0002          & 0.9457            & 0.0014           \\
\textbf{F19}   & 0.9800            & 0.0010            & 0.9586            & 0.0018           & 0.9686              & 0.0022             & 0.9229           & 0.0004          & 0.9957            & 0.0001           \\
\textbf{F20}   & 0.9771            & 0.0016            & 0.9757            & 0.0015           & 0.9800              & 0.0019             & 0.9686           & 0.0000          & 0.9857            & 0.0015           \\
\textbf{F21}   & 0.7514            & 0.0136            & 0.4414            & 0.0269           & 0.5371              & 0.0295             & 0.3214           & 0.0296          & 0.9143            & 0.0045           \\
\textbf{Avg}   & 0.9621            & 0.0022            & 0.9009            & 0.0056           & 0.9206              & 0.0045             & 0.8793           & 0.0068          & \textbf{0.9835}   & \textbf{0.0009}  \\ \hline
\end{tabular}
\end{adjustbox}

\end{table}

\begin{figure}[!ht]
	\centerline{\includegraphics[width=1.\columnwidth,height=0.5615\columnwidth]{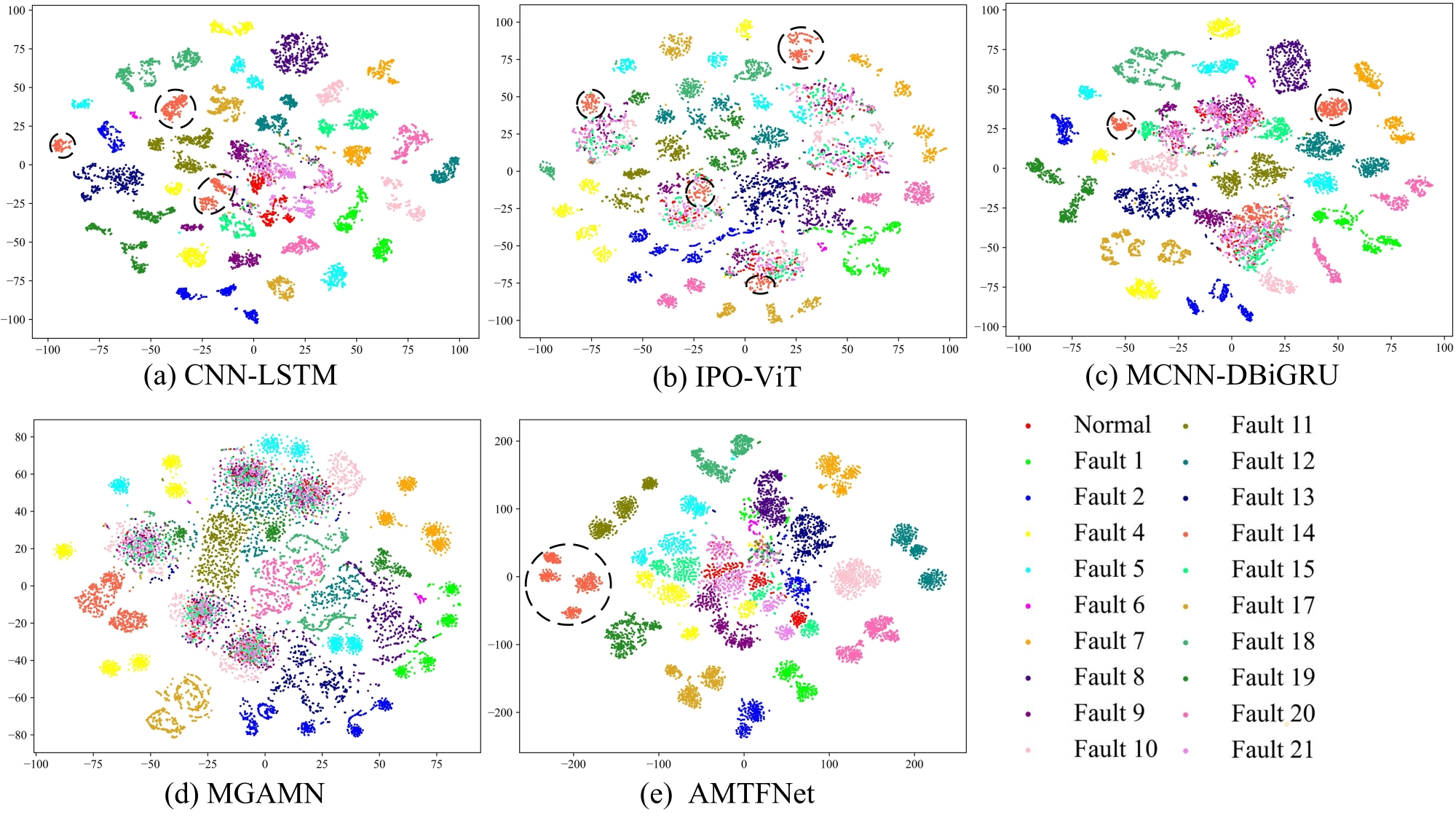}}
	\caption{t-SNE visualization results for task T1 on the TE process dataset.}
	\label{Fig8}
\end{figure}

The feature distributions learned by each model are visualized through t-SNE.
Taking task T1 as an example, the details of the visual representation are presented in \hyperref[Fig8]{Fig. 8}.
\hyperref[Fig8]{Fig. 8} (d) reveals that the features learned by MGAMN reveal poorly separated clusters.
Fault 14 was chosen for comparative analysis due to its consistent behavior under multiple operating modes and its representativeness among the faults.
The samples belonging to Fault 14 are highlighted with dashed boxes in \hyperref[Fig8]{Fig. 8} (a), (b), (c) and (e). 
It is evident  that the features extracted by CNN-LSTM, IPO-ViT and MCNN-DBiGRU are too scattered, whereas AMTFNet effectively clusters the samples belonging to this category.
This demonstrates that the features extracted by AMTFNet exhibit minimal differences across different modes, highlighting its superior cross-mode performance.

The experiment results for scalability comparison are presented in \hyperref[Table10]{Table 10}.
Models AMTFNet and MGAMN exhibit significantly fewer parameters than the other models, which makes them suitable for deployment in resource-constrained environments.
Furthermore, model AMTFNet demonstrates the shortest training time and relatively low test time.
The computational complexity of CNN-LSTM and IPO-ViT is significantly higher than that of AMTFNet, which highlights the advantages of AMTFNet in practical applications.

\begin{table}[!ht]
	\centering
    \begin{threeparttable}
	\caption{Experiment results for scalability comparison on the TE process dataset.}
    \small
	\label{Table10}
\begin{tabular}{cccccc}
\hline
\multirow{2}{*}{\textbf{Metrics}}                                           & \multicolumn{5}{c}{\textbf{Model}}                                                              \\ \cline{2-6} 
                                                                            & \textbf{CNN-LSTM} & \textbf{IPO-ViT} & \textbf{MCNN-DBiGRU} & \textbf{MGAMN} & \textbf{AMTFNet} \\ \hline
\textbf{\begin{tabular}[c]{@{}c@{}}Parameter \\ number (M)\end{tabular}}    & 0.91              & 26.39            & 0.51                 & 0.06           & 0.10             \\
\textbf{\begin{tabular}[c]{@{}c@{}}Training \\ time (s/epoch)\end{tabular}} & 11.61             & 114.20           & 10.77                & 7.85           & 7.41             \\
\textbf{\begin{tabular}[c]{@{}c@{}}Test time\\ (s/epoch)\end{tabular}}      & 3.19              & 6.35             & 2.67                 & 2.60           & 2.79             \\ \hline
\end{tabular}

\end{threeparttable}
\end{table}

\subsubsection{Ablation study}
To examine the capability of the MSDC, GRU, and TAM, ablation experiments were conducted. 
The Micro F1 scores of different ablation models are presented in \hyperref[Table11]{Table 11}.
The average Micro F1 score of A2 is 17.56\% higher than that of A1, indicating that the GRU is superior to the CNN in temporal feature extraction.
The model A3 and A4 are obtained by adding TAM to A1 and A2, respectively (see \hyperref[Table4]{Table 4}). 
The average Micro F1 score of A3 is 10.19\% higher than that of A1, and A4 is 1.63\% higher than that of A2. 
This demonstrates that TAM effectively fuses features extracted by both CNN and GRU.
A5 combines A1 and A2, meaning that both the MSDC and GRU are jointly employed to extract features. 
The average Micro F1 score of A5 is 20.35\% higher than that of A1 and 2.38\% higher than that of A2. 
This demonstrates that combining MSDC and GRU enhances the power to learn the complex features within the data.
Models A6 and AMTFNet are developed by integrating different attention mechanisms into the model A5. 
It is evident that the average Micro F1 score of A6 is lower than that of A5, whereas AMTFNet achieves a performance improvement.
These results indicate that the proposed TAM effectively enhances feature representations at critical moments.
The performance degradation of A6 may be attributed to the limitation of using only mean-based features, which are insufficient for effectively capturing critical features in complexly distributed data.
AMTFNet is equivalent to adding GRU to A3, or adding MSDC to A4, or adding TAM to A5. 
The average Micro F1 score of TAGRUN is 9.35\% higher than A3, 0.84\% higher than A4, and 0.11\% higher than A5. 
This indicates that GRU, MSDC, and TAM complement each other, collectively strengthening the model’s capability for uncertain-mode fault diagnosis.

\begin{table}[!ht]
	\centering
	\caption{Micro F1 scores of different ablation models on the TE process dataset.}

	\label{Table11}

    \begin{tabular}{lcllllll}
\hline
\multirow{2}{*}{\textbf{Task}} & \multicolumn{7}{c}{\textbf{Model}}                                                                                                                                                                                           \\ \cline{2-8} 
                               & \textbf{A1} & \multicolumn{1}{c}{\textbf{A2}} & \multicolumn{1}{c}{\textbf{A3}} & \multicolumn{1}{c}{\textbf{A4}} & \multicolumn{1}{c}{\textbf{A5}} & \multicolumn{1}{c}{\textbf{A6}} & \multicolumn{1}{c}{\textbf{AMTFNet}} \\ \hline
\textbf{T1}                    & 0.8118      & 0.9466                          & 0.9003                          & 0.9707                          & 0.9793                          & 0.9621                          & \textbf{0.9824}                      \\
\textbf{T2}                    & 0.8259      & 0.9595                          & 0.8889                          & 0.9743                          & 0.9748                          & 0.9642                          & \textbf{0.9758}                      \\
\textbf{T3}                    & 0.8058      & 0.9517                          & 0.8691                          & 0.9699                          & 0.9782                          & 0.9563                          & \textbf{0.9789}                      \\
\textbf{T4}                    & 0.8008      & 0.9612                          & 0.9259                          & 0.9743                          & 0.9808                          & 0.9695                          & \textbf{0.9814}                      \\
\textbf{T5}                    & 0.8292      & 0.9599                          & 0.9233                          & 0.9692                          & 0.9765                          & 0.9622                          & \textbf{0.9777}                      \\
\textbf{T6}                    & 0.8027      & 0.9533                          & 0.8657                          & 0.9676                          & \textbf{0.9790}                 & 0.9556                          & 0.9789                               \\
\textbf{Avg}                   & 0.8127      & 0.9554                          & 0.8955                          & 0.9710                          & 0.9781                          & 0.9617                          & \textbf{0.9792}                      \\ \hline
\end{tabular}

\end{table}

\begin{figure}[!ht]
	\centerline{\includegraphics[width=1.\columnwidth,height=0.361\columnwidth]{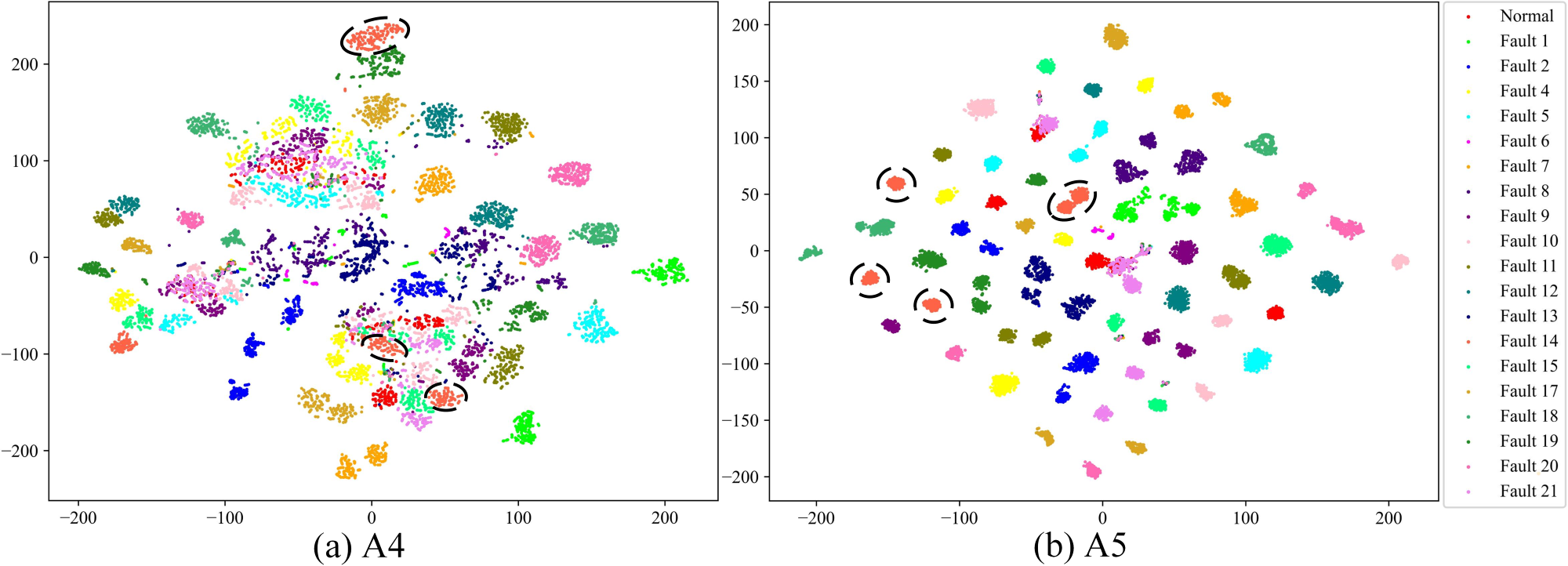}}
	\caption{t-SNE visualization results for ablation study on the TE process dataset.} 
	\label{Fig9}
\end{figure}

Since the improvement in AMTFNet's average Micro F1 score over A4 and A5 is relatively small, the feature maps extracted from A4 and A5 are visualized using t-SNE.
Taking Task T1 as an example, the visualization results are shown in  \hyperref[Fig9]{Fig. 9}.
Similarly, the samples belonging to Fault 14 are highlighted with dashed circles.
Although A4 achieves a Micro F1 score of 0.9707, the feature distributions of various classes exhibit poor separability. 
A5 effectively clusters the feature distributions of various classes into different groups. 
However, the features of samples within the same category exhibit significant dispersion.
This may be because A5 learns mode-specific features, causing each operation mode and system health condition category pair to be clustered into a unique group as tightly as possible.
For uncertain-mode fault diagnosis, it is essential to ensure both the separability of various categories and the compactness of the same category.
The proposed AMTFNet model achieves better intra-category compactness and maintains a reasonable degree of inter-category separability.
This demonstrates that focusing on critical time points through TAM may help the model capture invariant features across different modes, thereby reducing feature differences among samples of the same category.

\subsection{Case two: three-phase flow facility}
\subsubsection{Task description}
\begin{figure}[!ht]
	\centerline{\includegraphics[width=1.\columnwidth,height=0.4302\columnwidth]{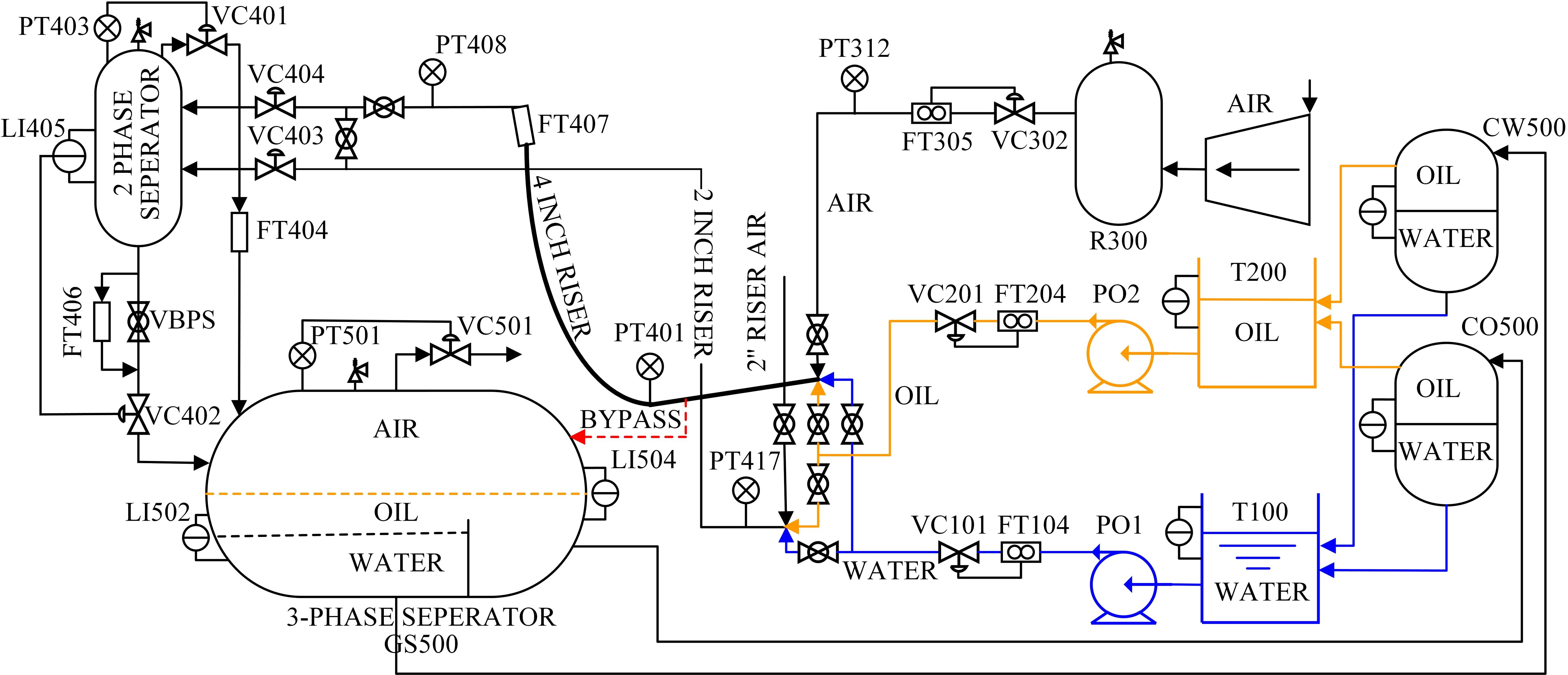}}
	\caption{Sketch of the TPFF \cite{RN260428}.}
	\label{Fig10}
\end{figure}

In the field of fault diagnosis, the dataset from three-phase flow facility (TPFF) \cite{RN260428} at Cranfield University has been commonly adopted to assess the effectiveness of various methods. 
The three-phase flow facility is designed to provide controlled flows of water, oil, and air.
Its structure is shown in \hyperref[Fig10]{Fig. 10}. 
This facility captured 24 process variables and simulated six faults.
The operation mode settings and the dataset used in the experiments are presented in \hyperref[Table12]{Table 12}, where different water flow rates and air flow rates correspond to different operating modes.
For each operating mode, the data collected from the initial time to the end of the fault event were used to construct the experimental dataset.

\begin{table}[!ht]
	\centering
	\caption{Faults of TPFF used in uncertain-mode fault diagnosis \cite{RN260428}.}
    \small
	\label{Table12}
\begin{tabular}{lll}
\hline
\textbf{Data set index} & \textbf{Water flow rate (kg/s)} & \textbf{Air flow rate ($\text{m}^3$/s)} \\ \hline
1.2, 4.2                & 2                               & 0.0417                                  \\
1.3, 3.3, 4.3           & 3.5                             & 0.0208                                  \\
2.2, 3.2                & 2                               & 0.0278                                  \\
2.3                     & 3.5                             & 0.0417                                  \\ \hline
\textbf{No.}            & \textbf{Description}            & \textbf{Data set index}                 \\ \hline
N                       & Normal                          & 1.2,1.3,2.2,2.3,3.2,3.3,4.2,4.3         \\
F1                      & Air line blockage               & 1.2,1.3                                 \\
F2                      & Water line blockage             & 2.2,2.3                                 \\
F3                      & Top separator input blockage    & 3.2,3.3                                 \\
F4                      & Open direct bypass              & 4.2,4.3                                 \\ \hline
\end{tabular}
\end{table}

\subsubsection{Experiment results}
The experiments for FDR, FPR, Micro F1 and Macro F1 scores are shown in \hyperref[Table13]{Table 13}.
AMTFNet obtains the highest FDR, average Micro F1, and average Macro F1 scores, all with values of 1.0.
In addition, the AMTFNet model achieved the lowest FPR, with a value of 0.
Compared to CNN-LSTM, IPO-ViT, MCNN-DBiGRU and MGAMN, AMTFNet's average FDR is higher by 0.01\%, 6.93\%, 1.08\% and 8.51\%, respectively; its average Micro F1 is higher by 0.03\%, 3.75\%, 0.63\% and 4.68\%, respectively; and its Macro F1 is higher by 0.03\%, 6.18\%, 0.98\% and 7.68\%, respectively.
For the system health condition category N, the FDR scores of IPO-ViT and MGAMN are all below 0.72, whereas both AMTFNet and CNN-LSTM achieve a score of 1. Additionally, CNN-LSTM achieves an FDR of 0.9993 for category F3, whereas AMTFNet still achieves a score of 1.
Overall, the AMTFNet model outperforms the other models across all four metrics, further validating its superiority in uncertain-mode fault diagnosis.

\begin{table}[!ht]
	\centering
    \begin{threeparttable}
	\caption{FDR, FPR, Micro F1  and Macro F1 scores of different methods on the TPFF dataset.}
    \small
	\label{Table13}
    \fontsize{10}{14}\selectfont
\begin{tabular}{llllcll}
\hline
\multirow{2}{*}{\textbf{Metrics}} & \multirow{2}{*}{\textbf{Class}} & \multicolumn{5}{c}{\textbf{Model}}                                                                                  \\ \cline{3-7} 
                                  &                                 & \textbf{CNN-LSTM} & \textbf{IPO-ViT} & \multicolumn{1}{l}{\textbf{MCNN-DBiGRU}} & \textbf{MGAMN} & \textbf{AMTFNet} \\ \hline
\textbf{FDR}                      & \textbf{N}                      & 1.0000            & 0.7135           & 0.9568                                   & 0.6324         & 1.0000           \\
                                  & \textbf{F1}                     & 1.0000            & 1.0000           & 1.0000                                   & 1.0000         & 1.0000           \\
                                  & \textbf{F2}                     & 1.0000            & 0.9653           & 0.9954                                   & 0.9838         & 1.0000           \\
                                  & \textbf{F3}                     & 0.9993            & 0.9972           & 0.9993                                   & 0.9917         & 1.0000           \\
                                  & \textbf{F4}                     & 1.0000            & 1.0000           & 0.9950                                   & 1.0000         & 1.0000           \\
                                  & \textbf{Avg}                    & 0.9999            & 0.9352           & 0.9893                                   & 0.9216         & \textbf{1.0000}  \\ \hline
\textbf{FPR}                      & \textbf{N}                      & 0.0003            & 0.0061           & 0.0019                                   & 0.0061         & 0.0000           \\
                                  & \textbf{F1}                     & 0.0000            & 0.0284           & 0.0004                                   & 0.0301         & 0.0000           \\
                                  & \textbf{F2}                     & 0.0000            & 0.0007           & 0.0000                                   & 0.0016         & 0.0000           \\
                                  & \textbf{F3}                     & 0.0000            & 0.0030           & 0.0000                                   & 0.0134         & 0.0000           \\
                                  & \textbf{F4}                     & 0.0000            & 0.0059           & 0.0052                                   & 0.0063         & 0.0000           \\
                                  & \textbf{Avg}                    & 0.0001            & 0.0088           & 0.0015                                   & 0.0115         & \textbf{0.0000}  \\ \hline
\textbf{$\text{F1}_{Micro}$}      & \textbf{}                       & 0.9997            & 0.9639           & 0.9937                                   & 0.9553         & \textbf{1.0000}  \\
\textbf{$\text{F1}_{Macro}$}      &                                 & 0.9997            & 0.9418           & 0.9903                                   & 0.9287         & \textbf{1.0000}  \\ \hline
\end{tabular}
\end{threeparttable}
\end{table}

\begin{figure}[!ht]
	\centerline{\includegraphics[width=1.\columnwidth,height=0.5575\columnwidth]{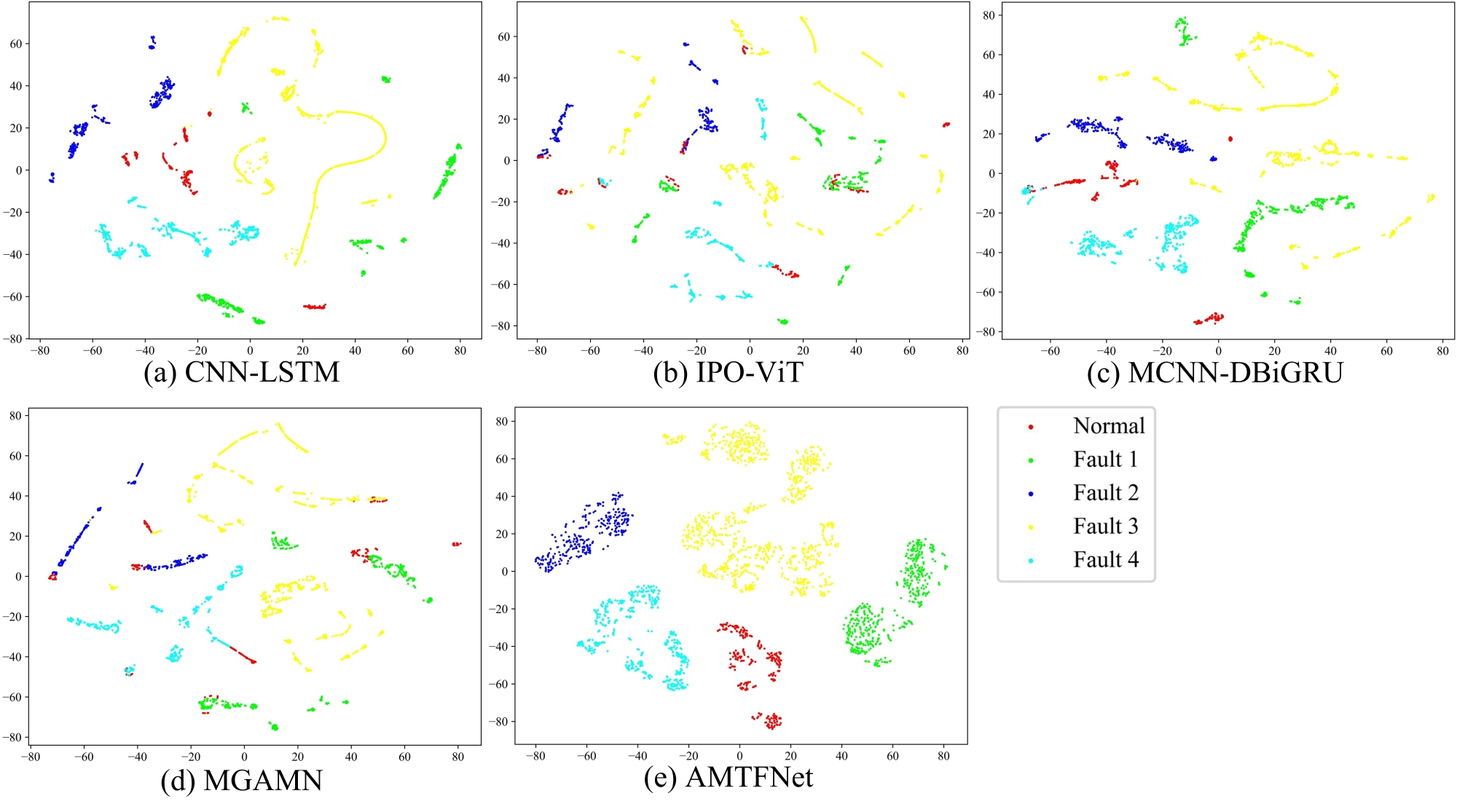}}
	\caption{t-SNE visualization results on the TPFF dataset.}
	\label{Fig11}
\end{figure}

The feature visualization results obtained using t-SNE are shown in \hyperref[Fig11]{Fig. 11}.
Although the average Micro F1 scores of CNN-LSTM and MCNN-DBiGRU are higher than 0.99, the comparison of \hyperref[Fig11]{Fig. 11}(a), (c) and (e) clearly shows that the features extracted by CNN-LSTM and MCNN-DBiGRU exhibit weaker cohesion of same-class and divergence among different classes compared to those extracted by AMTFNet.
The results shown in \hyperref[Fig11]{Fig. 11}(b) and (d) demonstrate that the features extracted by these models form poorly separable clusters.
In contrast, AMTFNet achieves promising results, with features of samples from the same class exhibiting regional distributions and clear inter-class distances.

The parameter number, training time per epoch, and testing time per epoch for each model are presented in \hyperref[Table14]{Table 14}. 
The results are consistent with those obtained in TE process. 
This demonstrates that AMTFNet is sufficiently lightweight, proving its advantage in practical applications.

\begin{table}[!ht]
	\centering
    \begin{threeparttable}
	\caption{Experiment results for scalability comparison on the TPFF dataset.}
    \small
	\label{Table14}
\begin{tabular}{cccccc}
\hline
\multirow{2}{*}{\textbf{Metrics}}                                           & \multicolumn{5}{c}{\textbf{Model}}                                                              \\ \cline{2-6} 
                                                                            & \textbf{CNN-LSTM} & \textbf{IPO-ViT} & \textbf{MCNN-DBiGRU} & \textbf{MGAMN} & \textbf{AMTFNet} \\ \hline
\textbf{\begin{tabular}[c]{@{}c@{}}Parameter \\ number (M)\end{tabular}}    & 0.69              & 26.35            & 0.50                 & 0.06           & 0.06             \\
\textbf{\begin{tabular}[c]{@{}c@{}}Training \\ time (s/epoch)\end{tabular}} & 4.26              & 42.73            & 4.52                 & 3.47           & 3.67             \\
\textbf{\begin{tabular}[c]{@{}c@{}}Test time\\ (s/epoch)\end{tabular}}      & 2.56              & 3.59             & 2.49                 & 2.27           & 2.45             \\ \hline
\end{tabular}

\end{threeparttable}
\end{table}

\section{Conclusion}
\label{Sec5}
This article proposed a model named AMTFNet for uncertain-mode fault diagnosis.
The AMTFNet stands out by focusing on key temporal moments that exhibit richer cross-mode information, significantly enhancing its ability to extract domain-invariant features. This improvement leads to superior performance in uncertain-mode fault diagnosis tasks.
The experiments on two datasets demonstrate that AMTFNet exhibits significant superiority in terms of fault diagnosis performance, visualization, and lightweight design.
The proposed method is capable of accurately assessing system health condition under various operating modes without the requirement of prior identification of the specific operating mode. 
This capability is particularly necessary in complex industrial systems with frequent mode transitions. 
In addition, the proposed method enables highly accurate state identification across different operating modes, thus facilitating timely intervention. 
Overall, this method significantly reduces the risk of process anomalies escalating into dangerous accidents and significantly enhances the safety and reliability of plant operations.

However, this study has certain limitations, particularly in that it does not consider fault scenarios arising during transitions between different operating modes.
The proposed fault diagnosis model is developed based on supervised learning, which relies on labeled data for training. 
Future research could explore the potential of unsupervised learning methods in uncertain-mode fault diagnosis. 
Additionally, the industrial systems may experience new faults or operate in new modes.
Therefore, exploring open-set uncertain-mode fault diagnosis and addressing the generalization problem in uncertain-mode fault diagnosis could be necessary future directions.




\bibliographystyle{elsarticle-num} 
\bibliography{ref}
\end{document}